\documentclass[a4paper, 10pt, twocolumn, journal, final]{ieeeconf}
\IEEEoverridecommandlockouts  
\usepackage[utf8]{inputenc}
\usepackage{graphicx}
\usepackage{silence}
\usepackage{caption}
\usepackage{epsfig} 
\usepackage{rotating}
\usepackage{fontawesome}  
\usepackage{mathptmx} 
\usepackage{times}  
\usepackage{amsmath} 
\usepackage{amssymb} 
\usepackage{dsfont} 
\usepackage{mathtools}
\usepackage{empheq}
\usepackage{eucal}
\usepackage{ragged2e}
\usepackage{bm} 
\usepackage{pgfgantt}
\usepackage{booktabs}  
\usepackage[flushleft]{threeparttable}
\usepackage[colorlinks=true, citecolor = blue, urlcolor = blue, linkcolor = blue]{hyperref}
\newcommand{\veclatin}[1]{\bm{#1}} 

\newcommand{\vecgreek}[1]{\pmb{#1}}

\usepackage{letterbib}
\usepackage{scalerel}
\usepackage[authoryear, round]{natbib}
\definecolor{barblue}{RGB}{153,204,254}
\definecolor{groupblue}{RGB}{51,102,254}
\definecolor{linkred}{RGB}{165,0,33}

\renewcommand\labelenumi{(\roman{enumi})}
\renewcommand\theenumi\labelenumi 

\usepackage{lmodern} 
\usepackage[nolists, tablesfirst]{endfloat}
\DeclareDelayedFloatFlavor{sidewaystable}{table}
\DeclareDelayedFloatFlavor{sidewaysfigure}{figure}

\renewcommand{\thefigure}{\arabic{figure}}
\renewcommand{\thetable}{\arabic{table}}
\usepackage{subcaption}

\title{\LARGE{\bf{Multi-Target XGBoostLSS Regression}}

\vspace{1em}

}
\author{\parbox{3 in}{\centering Alexander März$^{\aleph_{\scaleto{0\mathstrut}{2pt}}}$ 
        \thanks{$^{\aleph_{\scaleto{0\mathstrut}{4pt}}}$ \hspace{-0.8em} Author for correspondence: \href{mailto:alex.maerz@gmx.net}{alex.maerz@gmx.net}} \\ 
    }      
}

\begin{document}
\maketitle
\thispagestyle{empty}

\begin{abstract}
Current implementations of Gradient Boosting Machines are mostly designed for single-target regression tasks and commonly assume independence between responses when used in multivariate settings. As such, these models are not well suited if non-negligible dependencies exist between targets. To overcome this limitation, we present an extension of XGBoostLSS that models multiple targets and their dependencies in a probabilistic regression setting. Empirical results show that our approach outperforms existing GBMs with respect to runtime and compares well in terms of accuracy. 
\end{abstract}

\vspace{1em}

\section{Introduction}

The recent M5 forecasting competition demonstrated that tree-based models are highly competitive beyond cross-sectional tabular data.\footnote{For details on the M5 competition, see \citet{Makridakis.2021b, Makridakis.2021}. For a good overview of tree-based methods and their use in the M5 competition, see \cite{Januschowski.2021}.} Yet, despite their wide-spread use, when applied in a multivariate setting, current implementations of Gradient Boosting Machines (GBMs) commonly assume (conditional) independence between target variables.\footnote{While XGBoost \citep{Chen.2016} and CatBoost \citep{Prokhorenkova.2018, Dorogush.2017} allow modelling of several responses, with a separate model trained for each target, LightGBM \citep{Ke.2017} currently does not support multi-target regression. A workaround often suggested for extending models that do not natively support multi-target regression is to use scikit-learn's Multi-Output-Regressor. However, since a separate model is trained per target, this does not allow modelling of dependencies between multiple responses.} However, modelling of mutual dependencies in a probabilistic regression setting has been shown to increase accuracy and to also lead to added insight into the data generating process \citep{Schmid.2023}. Therefore, models tailored to single-target regression tasks are not well suited when applied in environments where non-negligible inter-target co-relations exist.\footnote{In the following, we use the terms multi-target and multivariate regression interchangeably for denoting environments where $\mathbf{y}$ is a $N \times D$ response matrix $\mathbf{y} = (y_{i1}, \ldots, y_{iD})^{T}, i=1, \ldots, N$ with $D$ denoting the target dimension.}

While high-dimensional multivariate forecasting is an active area of research in Deep Learning (see, e.g., \cite{Kan.2022, Rasul.2021b, Rasul.2021, Wu.2020, Salinas.2019}) with applications ranging from anomaly detection to causal analysis or retail sales forecasting, where modelling of dependencies between articles is crucial to account for cannibalization effects, tree-based approaches have received comparatively few multivariate extensions. Recent advances include \cite{Pande.2022} who introduce a gradient boosting approach to model multivariate longitudinal responses, as well as \cite{Nespoli.2022} who present a computationally efficient algorithm for fitting multivariate boosted trees. To address the problem of low generalization ability and tree redundancy when dependencies between several targets are ignored, \cite{Zhang.2021} propose a general method to train GBMs with multiple targets. \cite{OMalley.2021} extend the NGBoost model of \cite{Duan.2020} to a multivariate Gaussian setting and \cite{Cevid.2021} introduce a distributional regression forest for multivariate responses. Based on Bayesian additive regression trees (BARTs) of \cite{Chipman.2010}, \cite{Clark.2021} develop multivariate time series models and \cite{Um.2020} extend BARTs to allow modelling of multivariate skewed responses. \cite{Lang.2020} introduce multivariate distributional regression forests to probabilistically predict wind profiles. \cite{Quan.2018} use multivariate trees to model insurance claims data with correlated responses and \cite{Miller.2016} introduce a multivariate extension of gradient boosted regression trees for continuous multivariate responses. 

With this paper, we contribute to the emerging literature on multi-target probabilistic GBMs and present a multivariate extension of the univariate XGBoostLSS introduced by \cite{Marz.2019}. Our approach leverages automatic differentiation using PyTorch \citep{Paszke.2019}, which facilitates implementation of distributions for which Gradients and Hessians are difficult to derive analytically.

The remainder of this paper is organized as follows: Section \ref{sec:mXGBoostLSS} introduces our multivariate XGBoostLSS framework and Section \ref{sec:applications} presents both a simulation study and real world examples. Section \ref{sec:conclusion} concludes.\footnote{The code of the implementation will be made available on \faGithub\href{https://github.com/StatMixedML/XGBoostLSS}{StatMixedML/XGBoostLSS} at the time of the final publication of the paper.}


\section{Multi-Target XGBoostLSS} \label{sec:mXGBoostLSS}

In its original formulation, distributional modelling relates all parameters of a univariate response distribution to covariates $\mathbf{x}$. In particular, it assumes the  response to follow a distribution $\mathcal{D}\bigl(\veclatin{\theta}(\mathbf{x})\bigr)$ that depends on up to four parameters, i.e., $y_{i} \stackrel{ind}{\sim} \mathcal{D}(\mu_{i\mathbf{x}}, \sigma^{2}_{i\mathbf{x}}, \nu_{i\mathbf{x}}, \tau_{i\mathbf{x}}), i=1,\ldots,N$, where $\mu_{i\mathbf{x}}$ and $\sigma^{2}_{i\mathbf{x}}$ are often location and scale parameters, respectively, while $\nu_{i\mathbf{x}}$ and $\tau_{i\mathbf{x}}$ correspond to shape parameters such as skewness and kurtosis.\footnote{While the original formulation of GAMLSS in \citet{Rigby.2005} suggests that any distribution can be described by location, scale and shape parameters, it is not necessarily true that the observed data distribution can actually be characterized by all of these parameters. Hence, we prefer to use the term distributional modelling.} More generally, univariate distributional modelling can be formulated as follows

\begin{empheq}[left=y_{i} \stackrel{ind}{\sim} \mathcal{D} \empheqbiglparen, right=\empheqbigrparen]{align}
	h_{1}\big(\theta_{i1}(\mathbf{x}_{i})\big) &= \eta_{i1} \nonumber\\ 
	h_{2}\big(\theta_{i2}(\mathbf{x}_{i})\big) &= \eta_{i2}  \label{eq:dist_model}   \nonumber \\  
	\vdots \nonumber \\                        
	h_{K}\big(\theta_{iK}(\mathbf{x}_{i})\big) &= \eta_{iK} \nonumber 
\end{empheq}

\noindent for $i = 1, \ldots, N$, where $\mathcal{D}(\cdot)$ denotes a parametric distribution for the response $\textbf{y} = (y_{1}, \ldots, y_{N})^{T}$ that depends on $K$ distributional parameters $\theta_{k}$, $k = 1, \ldots, K$, and with $h_{k}(\cdot)$ denoting a known function relating distributional parameters to predictors $\vecgreek{\eta}_{k}$. The predictor specification $\vecgreek{\eta}_{k} = f_{k}(\mathbf{x}), k = 1, \ldots, K$ is generic enough to use either GAM-type, Deep Learning or GBMs as in our case. 

To allow for a more flexible framework that explicitly models the dependencies of a $D$-dimensional response $\mathbf{y} = (y_{i1}, \ldots, y_{iD})^{T}, i=1, \ldots, N$, \cite{Klein.2015} introduce a multivariate version of distributional regression. Similar to the univariate case, multivariate distributional regression relates all $K$ parameters of a multivariate density $f_{i}\big(y_{i1}, \ldots, y_{iD} | \theta_{i1}(\mathbf{x}\big), \ldots, \theta_{iK}(\mathbf{x})\big)$ to a set of covariates $\mathbf{x}$.\footnote{To improve on the convergence and stability of XGBoostLSS estimation, unconditional Maximum Likelihood estimates of the parameters $\theta_{k}$, $k = 1, \ldots, K$ are used as offset values. Also, since XGBoostLSS updates the parameters by optimizing Gradients and Hessians, it is important that these are comparable in magnitude for all distributional parameters. Due to variability regarding the ranges, the estimation of Gradients and Hessians might become unstable so that XGBoostLSS might not converge or might converge very slowly. To mitigate these effects, we have implemented a stabilization of Gradients and Hessians.}

\subsection{Multivariate Gaussian Regression} \label{sec:mvn}

A common choice for multivariate probabilistic regression is to assume a multivariate Gaussian distribution, with the density given as \footnote{In the further course of this section, we follow the notation of \citet{Muschinski.2022, OMalley.2021, Salinas.2019}.}

\begin{equation}
	f\big(\mathbf{y}|\vecgreek{\theta}_{\mathbf{x}}\big) = \frac{1}{\sqrt{(2\pi)^{D}|\vecgreek{\Sigma}_{\mathbf{x}}|}}\exp\left(-\frac{1}{2}(\mathbf{y} - \vecgreek{\mu}_{\mathbf{x}})^{T} \vecgreek{\Sigma}^{-1}_{\mathbf{x}} (\mathbf{y} - \vecgreek{\mu}_{\mathbf{x}})\right) \nonumber
\end{equation}

\noindent where $\vecgreek{\mu}_{\mathbf{x}} \in \mathbb{R}^{D}$ represents a vector of conditional means, $\vecgreek{\Sigma}_{\mathbf{x}}$ is a positive definite symmetric $D \times D$ covariance matrix and $|\cdot|$ denotes the determinant. For the bivariate case $D=2$, $\vecgreek{\Sigma}_{\mathbf{x}}$ can be expressed as 

\begin{equation}
	\vecgreek{\Sigma}_{i\mathbf{x}} = \begin{bmatrix}
	\sigma^{2}_{i,1}(\mathbf{x}) & \rho_{i}(\mathbf{x})\sigma_{i,1}(\mathbf{x})\sigma_{i,2}(\mathbf{x})  \\
	\rho_{i}(\mathbf{x})\sigma_{i,2}(\mathbf{x})\sigma_{i,1}(\mathbf{x}) & \sigma^{2}_{i,2}(\mathbf{x})  \\
	\end{bmatrix} \nonumber
\end{equation}

\noindent with the variances on the diagonal and the covariances on the off-diagonal, for $ i=1, \ldots, N $. 

\subsection*{\hfil Cholesky Decomposition of Covariance Matrix \hfil} \label{sec:cholesky}

To ensure positive definiteness of $\vecgreek{\Sigma}$, the $D(D+1)/2$ entries of the covariance matrix must satisfy specific conditions.\footnote{Without loss of generality, we notationally omit the explicit dependency of all parameters on $\mathbf{x}$ and $i=1,\ldots,N$ in the following.} For the bivariate case, this can be ensured by applying exponential functions to the variances and a suitable transformation to restrict the coefficient of correlation $\rho \in [-1,1]$. However, in high-dimensional settings, where all moments are modelled as functions of covariates, ensuring positive definiteness of the covariance matrix becomes challenging, since joint restrictions for the elements of $\vecgreek{\Sigma}$ are necessary \citep{Muschinski.2022}. A computationally more tractable approach to ensure positive definiteness is based on the Cholesky-decomposition, that uniquely decomposes the covariance matrix as follows

\begin{equation}
	\vecgreek{\Sigma} = \mathbf{L}\mathbf{L}^{T} \nonumber 	
\end{equation}

\noindent where $\mathbf{L} \in \mathbb{R}^{D \times D}$ is a lower triangular matrix.\footnote{For the precision matrix, the Cholesky-decomposition as given as $\vecgreek{\Sigma}^{-1} = (\mathbf{L}^{-1})^{T}\mathbf{L}^{-1}$.} To ensure $\vecgreek{\Sigma}$ to be positive definite, the $D$ diagonal elements $\ell_{ii}$ of $\mathbf{L}$ need to be strictly positive, whereas all $D(D-1)/2$ off diagonal elements $\ell_{ij}$ can take on any value in $\mathbb{R}$, leaving them untransformed. Illustrative for the bivariate case, the Cholesky factor $\mathbf{L}$ is given as follows\footnote{In contrast to the original formulation of $\vecgreek{\Sigma}$, the elements in $\mathbf{L}$ do not have any direct interpretation.}

\begin{equation}
	\mathbf{L} = \begin{bmatrix}
	\exp(\ell_{11}) &  0  \\
	\ell_{21} & \exp(\ell_{22}) \\
	\end{bmatrix} \nonumber
\end{equation}

\noindent In addition to reparameterizing the covariance matrix, the Cholesky decomposition is also computationally efficient, since only the determinant of a triangular matrix needs to be calculated \citep{Salinas.2019}.

\subsection*{\hfil Low-Rank Covariance Approximation \hfil} \label{sec:lra_reg}

While efficient for low to medium dimensions of $D$, the computational cost of the Cholesky-decomposition becomes prohibitive in high-dimensional settings. To reduce the computational overhead, the covariance matrix $\vecgreek{\Sigma}$ can be approximated via the sum of a diagonal matrix $\mathbf{K} \in \mathbb{R}^{D \times D}_{+}$ and a unrestricted low-rank matrix $\mathbf{V} \in \mathbb{R}^{D \times r}$ 

\vspace{-1em}

\begin{align}
	\vecgreek{\Sigma} &= \mathbf{K} + \mathbf{V}\mathbf{V}^{T}  \nonumber \\
	                  &= \begin{bmatrix}
	                   \exp(\text{K}_{1}) & \dots  & 0\\
	                  \vdots & \ddots & \vdots\\
	                  0 & \dots  & \exp(\text{K}_{D})
	                  \end{bmatrix} + 
	                  \begin{bmatrix}
	                   \text{V}_{1}  \\
	                   \vdots  \\
	                   \text{V}_{D} \\
	                   \end{bmatrix} 
	                  \begin{bmatrix}
	                   \text{V}_{1}  \\
	                   \vdots  \\
	                   \text{V}_{D} \\
	                  \end{bmatrix}^{T} \nonumber
\end{align}

\noindent where $\exp(\cdot)$ ensures all diagonal entries of $\mathbf{K}$ to be strictly positive and the rank parameter $r$ governs the quality of the approximation. The computational efficiency of this approach results from the fact that the rank parameter $r<\!\!<D$ can typically be chosen much smaller than the number of target variables $D$ \citep{Salinas.2019}. Showing the relationship between the response dimension $D$ and the number of parameters $K$, Table \ref{tab:n_params} indicates that the number of parameters increases exponentially for the Cholesky-decomposition, while the relationship is only linear for the low-rank approximation, making it more suitable for high-dimensional settings.

\begin{table}[h!]
	\begin{center}
			\begin{threeparttable}
				\caption{Number of parameters for Cholesky and Low-Rank Approximation (LRA)}
				\begin{tabular}{lccccc}
					\toprule
					$\mbox{Y}_{D}$ & Cholesky & LRA(r=5) & LRA(r=10) & LRA(r=20) \\
					\midrule
					2 & 5 & 14 & 24 & 44 \\
					5 & 20 & 35 & 60 & 110 \\
					10 & 65 & 70 & 120 & 220 \\
					50 & 1,325 & 350 & 600 & 1,100 \\
					100 & 5,150 & 700 & 1,200 & 2,200 \\
					500 & 125,750 & 3,500 & 6,000 & 11,000 \\
					1,000 & 501,500 & 7,000 & 12,000 & 22,000 \\
					10,000 & 50,015,000 & 70,000 & 120,000 & 220,000 \\
					\bottomrule
				\end{tabular}
				\begin{tablenotes}
					\scriptsize
					\item \hspace{-0.7em} The table shows the number of parameters $K$ to estimate for a Multivariate Gaussian for the Cholesky $D(D+3)/2$ and the low-rank covariance matrix approximation $D(2+r)$ as functions of the response dimension $\mbox{Y}_{D}$. 
				\end{tablenotes}
				\label{tab:n_params}
		\end{threeparttable}
	\end{center}
\end{table}

\subsection{Multivariate Student-T Regression} \label{sec:student}

As a generalization of the multivariate Gaussian, the multivariate Student-T is suitable when modelling heavy-tailed data, i.e., when there is more mass in the tails of the distribution. The density is given as

\begin{equation}
	\begin{array}{l}
		f\big(\mathbf{y}|\vecgreek{\theta}_{\mathbf{x}}\big) = \nonumber \vspace{1em} \\ 		
		{\frac {\Gamma \left[\frac{\vecgreek{\nu}_{\mathbf{x}} + D}{2}\right]}{\Gamma\left[\frac{\vecgreek{\nu}_{\mathbf{x}}}{2}\right](\pi\vecgreek{\nu}_{\mathbf{x}})^{D/2}\left|{\boldsymbol {\Sigma_{\mathbf{x}} }}\right|^{1/2}}}\left[1+{\frac{({\mathbf{y}}-{\boldsymbol {\mu}_{\mathbf{x}}})^{T}{\boldsymbol {\Sigma }}^{-1}_{\mathbf{x}}({\mathbf {y} }-{\boldsymbol{\mu}_{\mathbf{x}}})}{\vecgreek{\nu_{\mathbf{x}}}}}\right]^{-\frac{\vecgreek{\nu}_{\mathbf{x}} + D}{2}} \nonumber
	 \end{array}
\end{equation}

\noindent with covariance matrix $\vecgreek{\nu}_{\mathbf{x}}(\vecgreek{\nu}_{\mathbf{x}}-2)^{-1}\vecgreek{\Sigma}_{\mathbf{x}}$ and $\Gamma\left[\cdot\right]$ denoting the gamma function. $\vecgreek{\mu}_{\mathbf{x}} \in \mathbb{R}^{D}$ and $\vecgreek{\Sigma}_{\mathbf{x}}$ are defined as for the multivariate Gaussian. The multivariate Student-T distribution has an additional degrees of freedom parameter $\vecgreek{\nu}_{\mathbf{x}} > 2$ that governs the tail behaviour, where for $\vecgreek{\nu}_{\mathbf{x}} \rightarrow \infty$ the Student-T converges in distribution to the multivariate Normal. Similar to the multivariate Gaussian, we use the Cholesky-decomposition $\vecgreek{\Sigma}_{\mathbf{x}} = \mathbf{L}_{\mathbf{x}}\mathbf{L}^{T}_{\mathbf{x}}$ to ensure the covariance matrix to be positive definite.\footnote{Unlike the multivariate Normal, a low-rank approximation of the covariance matrix is not yet implemented for the multivariate Student-T, but is planned in future releases.}

\subsection{Dirichlet Regression} \label{sec:dirichlet}

While the multivariate Gaussian and Student-T are defined for $\mathbf{y} \in \mathbb{R}^{D}$, the Dirichlet distribution is commonly used for modelling non-negative compositional data, i.e., data that consist of sub-sets that are fractions of some total. Compositional data are typically represented as proportions or percentages summing to 100\%, so that the Dirichlet extends the univariate beta-distribution to the multivariate case \citep{Klein.2015}. Dating back to the seminal paper of \cite{Aitchison.1982}, compositional data analysis (CoDa) is a branch of statistics that deals with multivariate observations carrying relative information and finds widespread use in ecology \citep{Douma.2019}, economics \cite{Fry.2000} or political science \citep{Katz.1999}. As a result of the unit-sum constraint, models that use distributions designed for unconstrained data typically suffer from the problem of spurious correlation when applied to compositional data \citep{Aitchison.2003}. 

The density of the Dirichlet distribution with parameters $\veclatin{\alpha}_{\mathbf{x}} = (\alpha_{\mathbf{x},1}, \ldots, \alpha_{\mathbf{x},D}) \in \mathbb{R}^{D}_{+}$ with $\sum^{D}_{d=1}y_{d}=1$ for all $y_{d}\in \left[0,1\right]$ is given by

\begin{equation}
	f\big(\mathbf{y}|\vecgreek{\theta}_{\mathbf{x}}\big) = \frac{1}{\mathrm{B}(\veclatin{\alpha}_{\mathbf{x}})} \prod_{d=1}^{D}y^{\alpha_{\mathbf{x},d-1}}_{d} \nonumber
\end{equation}

\noindent where the normalizing constant is expressed as the multinomial beta-function

\begin{equation}
	\mathrm{B}(\boldsymbol{\alpha}_{\mathbf{x}}) = {\frac{\prod\limits_{d=1}^{D}\Gamma(\alpha_{\mathbf{x},d})}{\Gamma\left(\sum \limits _{d=1}^{D}\alpha_{\mathbf{x},d}\right)}} \nonumber
\end{equation}

\noindent To ensure positivity, we use $\exp(\alpha_{\mathbf{x},d})$ for all $d=1,\ldots, D$. The estimated parameters have the interpretation of providing the probability of an event falling into category $d$, i.e., $\mathbb{E}(y_{d}) = \frac{\alpha_{d}}{\alpha_{0}}$, with $\alpha_{0} = \sum^{D}_{d=1}\alpha_{d}$ \citep{Klein.2015}.

\section{Applications} \label{sec:applications}

In this section, we present simulation studies and real-world examples to illustrate the functionality of our approach. All hyper-parameters of the models presented in this paper are selected using Optuna of \cite{Akiba.2019}. For all models, Table \ref{tab:hyperparams} shows the space of the tuneable hyper-parameter search.

\begin{table}[h!]
	\begin{center}
		\begin{threeparttable}
			\caption{Hyper-Parameter Search-Space}
			\begin{tabular}{lr}
				\toprule
				& Range \\ 
				\midrule
				learning-rate           & [0.001, 1.0]  \\
				max-depth               & [2, 10]       \\                        
				gamma                   & [0, 100]      \\  
				sub-sample              & [0.4, 1.0]     \\  
				col-sample              & [0.4, 1.0]     \\  
				min-child-weight        & [0, 500]      \\  
				boosting-iterations     & [500]       \\  
				early-stopping-rounds   & [2]     \\  
				\bottomrule
			\end{tabular}
			\label{tab:hyperparams} 
		\end{threeparttable}
	\end{center}
\end{table}

\subsection{Simulation} \label{sec:simulation} 

\subsection*{\hfil Multivariate Gaussian Regression \hfil}

We start with a trivariate Gaussian scenario ($N=10,000$), where all moments of the distribution are allowed to vary with covariates $\mathbf{x}$.\footnote{Hyper-parameters for each of the models in the simulation study are optimized running 100 hyper-parameter trails each.}

\begin{figure}[h!]
	\centering
	\caption{Estimated Parameters of Trivariate Gaussian.}
	\begin{subfigure}{1.0\textwidth}
		\centering
		\caption{Estimated Parameters using Cholesky-Decomposition of Covariance-Matrix.}
		\includegraphics[width=1.0\linewidth]{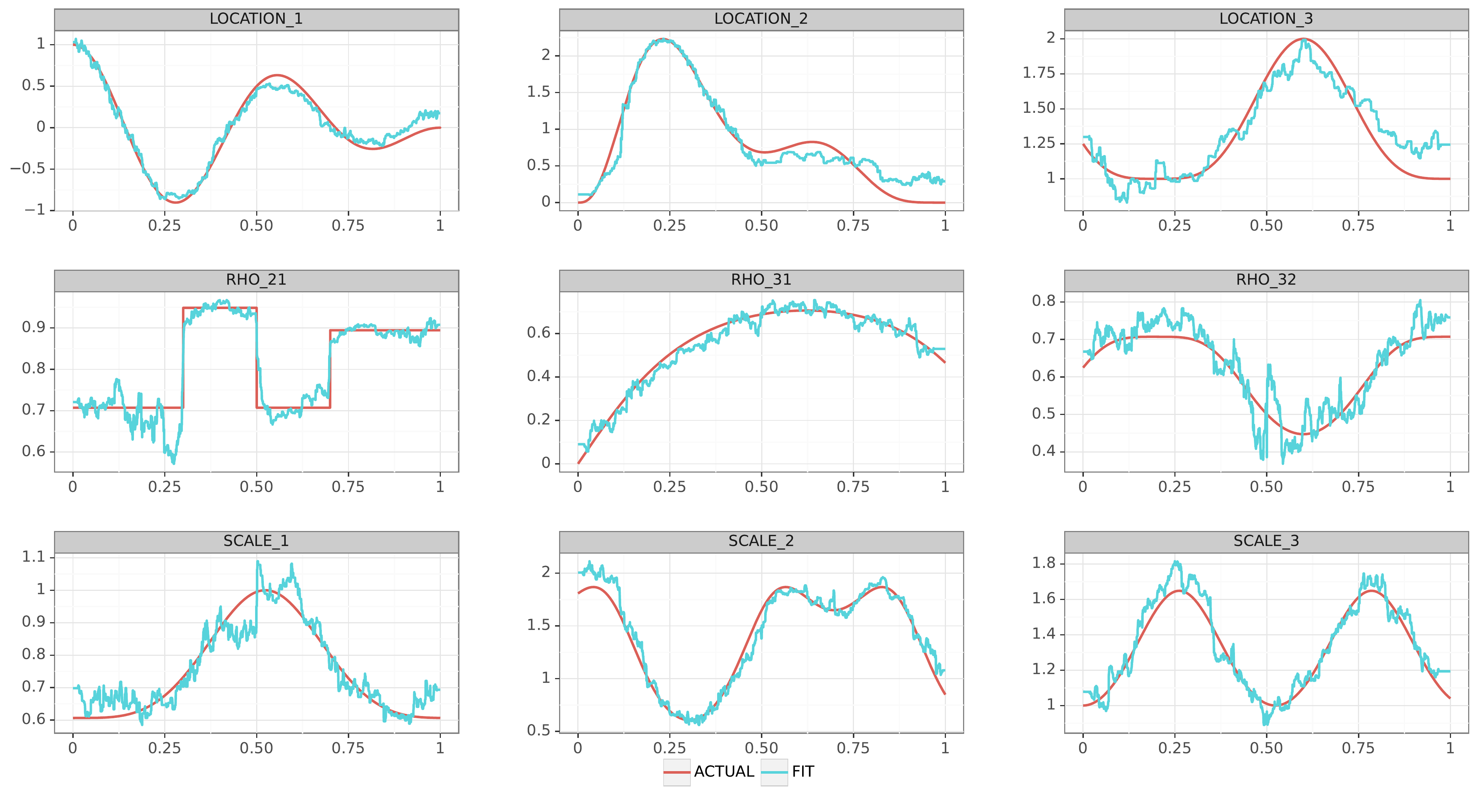}
		\label{fig:sim_mvn_cholesky}
	\end{subfigure}

\vspace{2em}

	\begin{subfigure}{1.0\textwidth}
		\centering
		\caption{Estimated Parameters using a Low-Rank-Approximation ($r=2$) of Covariance-Matrix.}
		\includegraphics[width=1.0\linewidth]{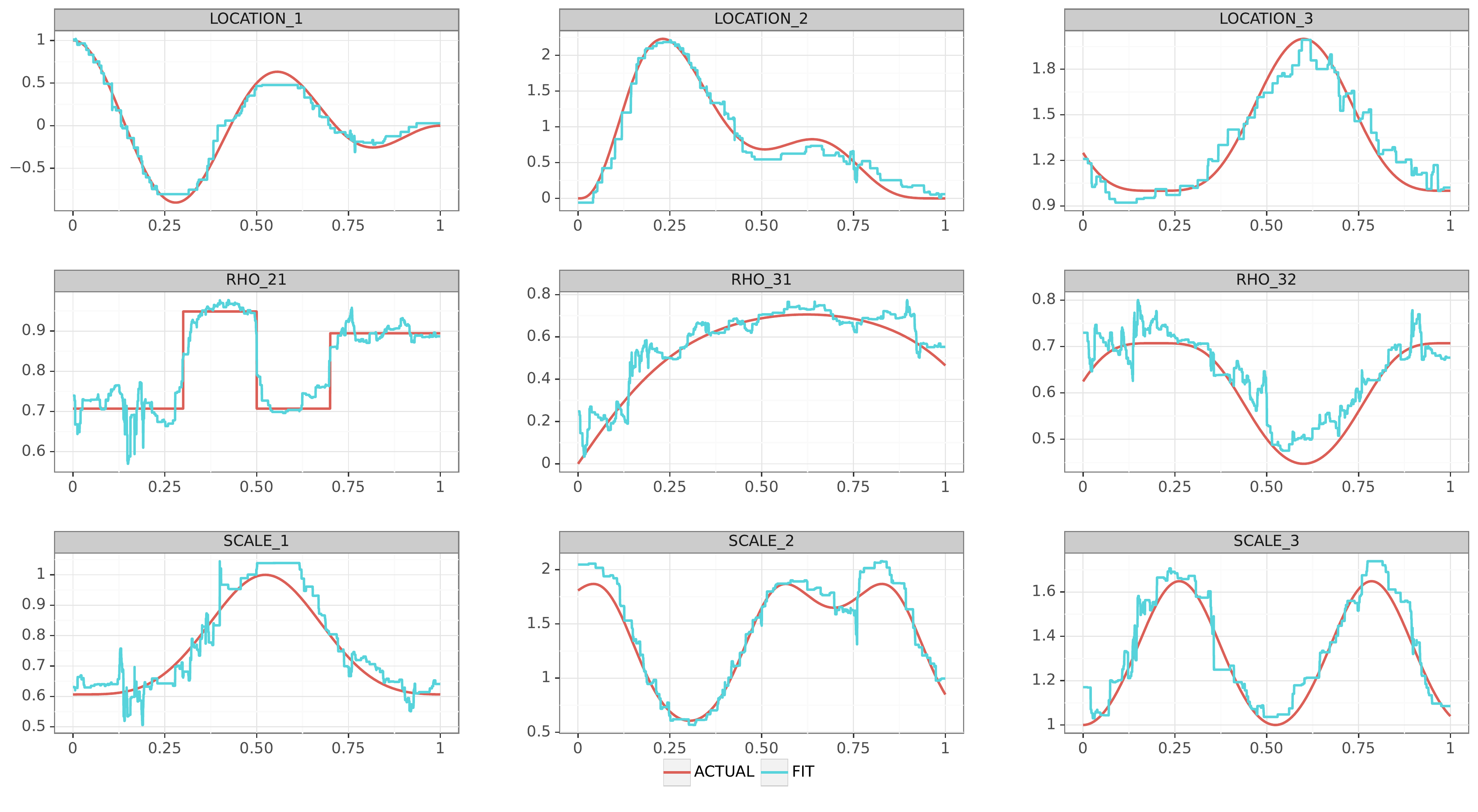}
		\label{fig:sim_mvn_lra}
	\end{subfigure}
	\label{fig:sim_mvn}
\end{figure}

\noindent Figure \ref{fig:sim_mvn} shows that the estimated parameters of the multivariate Gaussian, with the covariance matrix being parametrized using either the Cholesky-decomposition (Panel \ref{fig:sim_mvn_cholesky}) or the low-rank approximation (Panel \ref{fig:sim_mvn_lra}), closely match the true parameters, even for a rank as low as 2. Yet, for $\veclatin{\rho}_{21}$ and  $\veclatin{\rho}_{32}$ specifically, the low-rank approximation shows some deviations and generally a somewhat more erratic behaviour of the estimates as compared to the Cholesky-decomposition. 

\subsection*{\hfil Multivariate Student-T Regression \hfil}

\begin{figure}[h!]
	\centering
	\caption{Estimated Parameters of trivariate Student-T.}
	\includegraphics[width=1.0\linewidth]{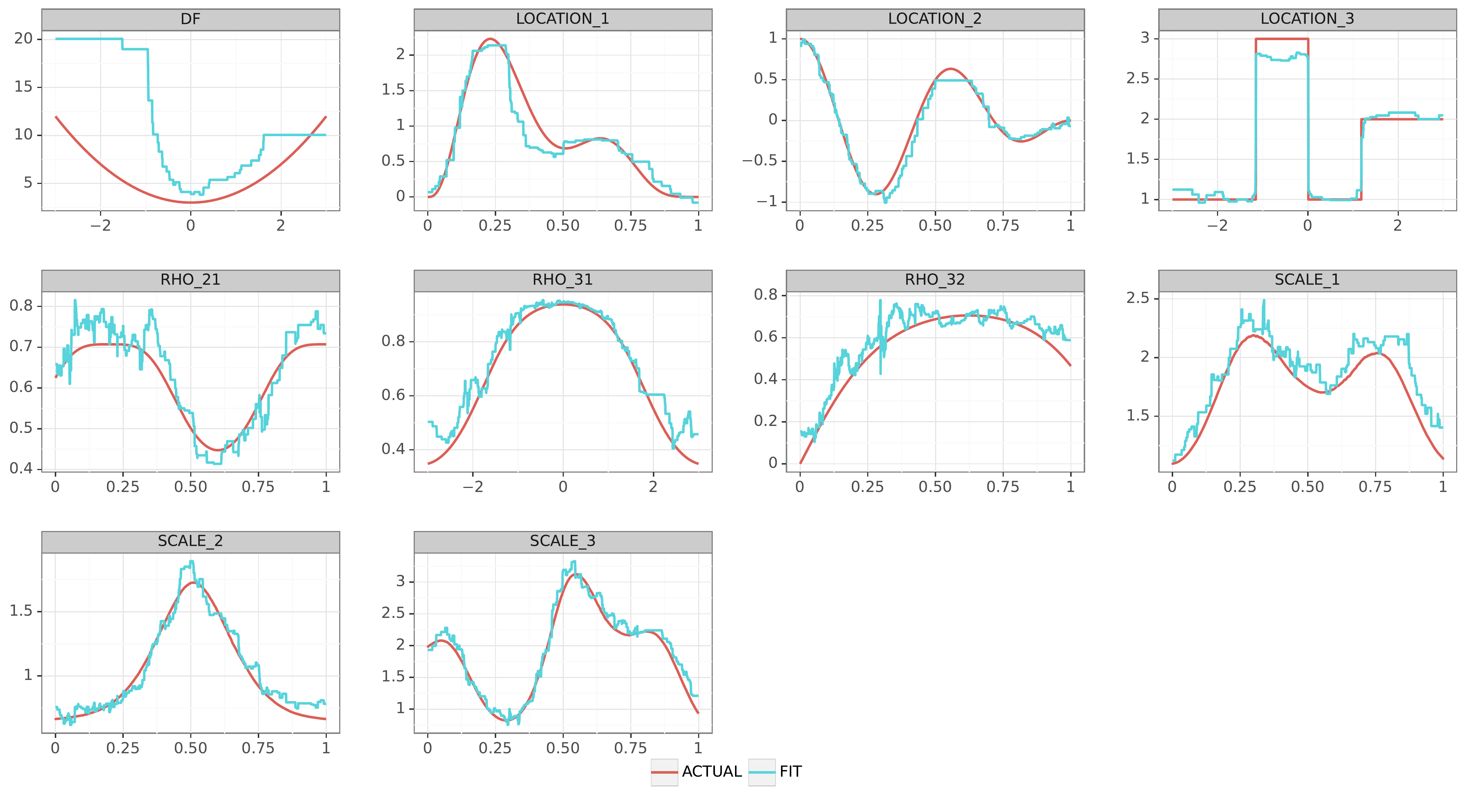}
	\label{fig:sim_mvt_cholesky}
\end{figure}

Figure \ref{fig:sim_mvt_cholesky} presents the simulation results for the trivariate Student-T distribution, where the covariance matrix is parametrized via the estimated Cholesky-factors $\hat{\mathbf{L}}\hat{\mathbf{L}}^{T}$.\footnote{Similar to the Gaussian scenario, we set $N=10,000$.}

\noindent The estimates presented in Figure \ref{fig:sim_mvt_cholesky} are still quite close to the true shapes, but not as accurate as for the multivariate Gaussian. In general, the Student-T estimates exhibit a somewhat more erratic behaviour than the Gaussian Cholesky-decomposition. This is especially true for the degrees of freedom parameter $\veclatin{\nu}$, which does not approximate the U-shape well. Also, parameter estimates of $\veclatin{\mu}_{3}$ and $\veclatin{\rho}_{21}$ deviate slightly from the true values.

\subsection*{\hfil Dirichlet Regression \hfil}

\begin{figure}[h!]
	\centering
	\caption{Relative Frequencies of Sand, Silt, and Clay in Arctic-Lake Data.}
	\includegraphics[width=0.36\linewidth]{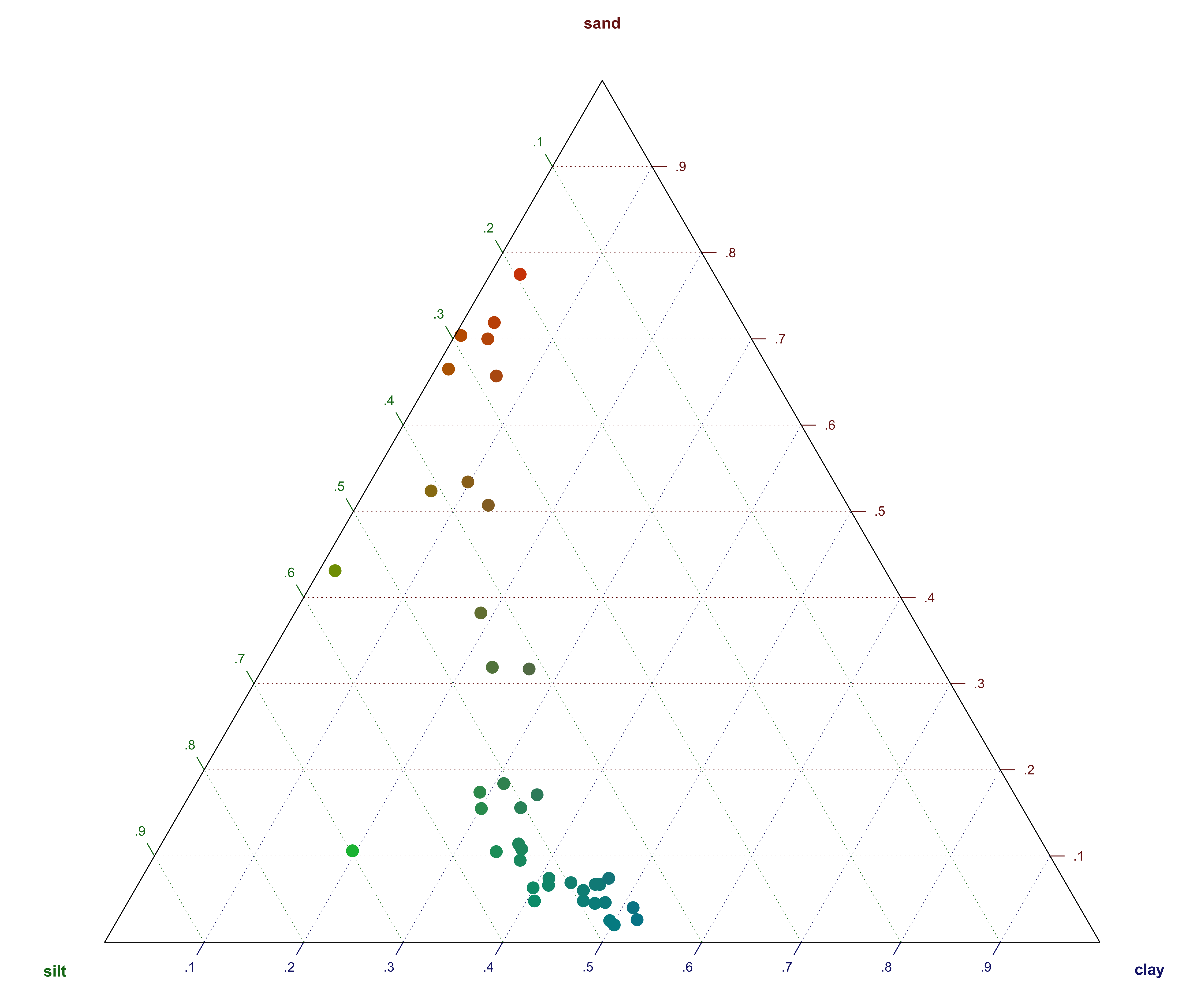}
	\label{fig:arctic_lake_composition}
\end{figure}

To evaluate the ability of our approach of inferring the relationship between covariates and a set of Dirichlet-distributed responses, we apply our model to the widely used Arctic-Lake dataset of \citep{Aitchison.2003} that contains information on sediment composition (sand, silt, clay) of an Arctic lake. The data are shown in Figure \ref{fig:arctic_lake_composition}.

\begin{figure}[h!]
	\centering
	\caption{Sediment Composition of Arctic-Lake Data.}
	\includegraphics[width=0.45\linewidth]{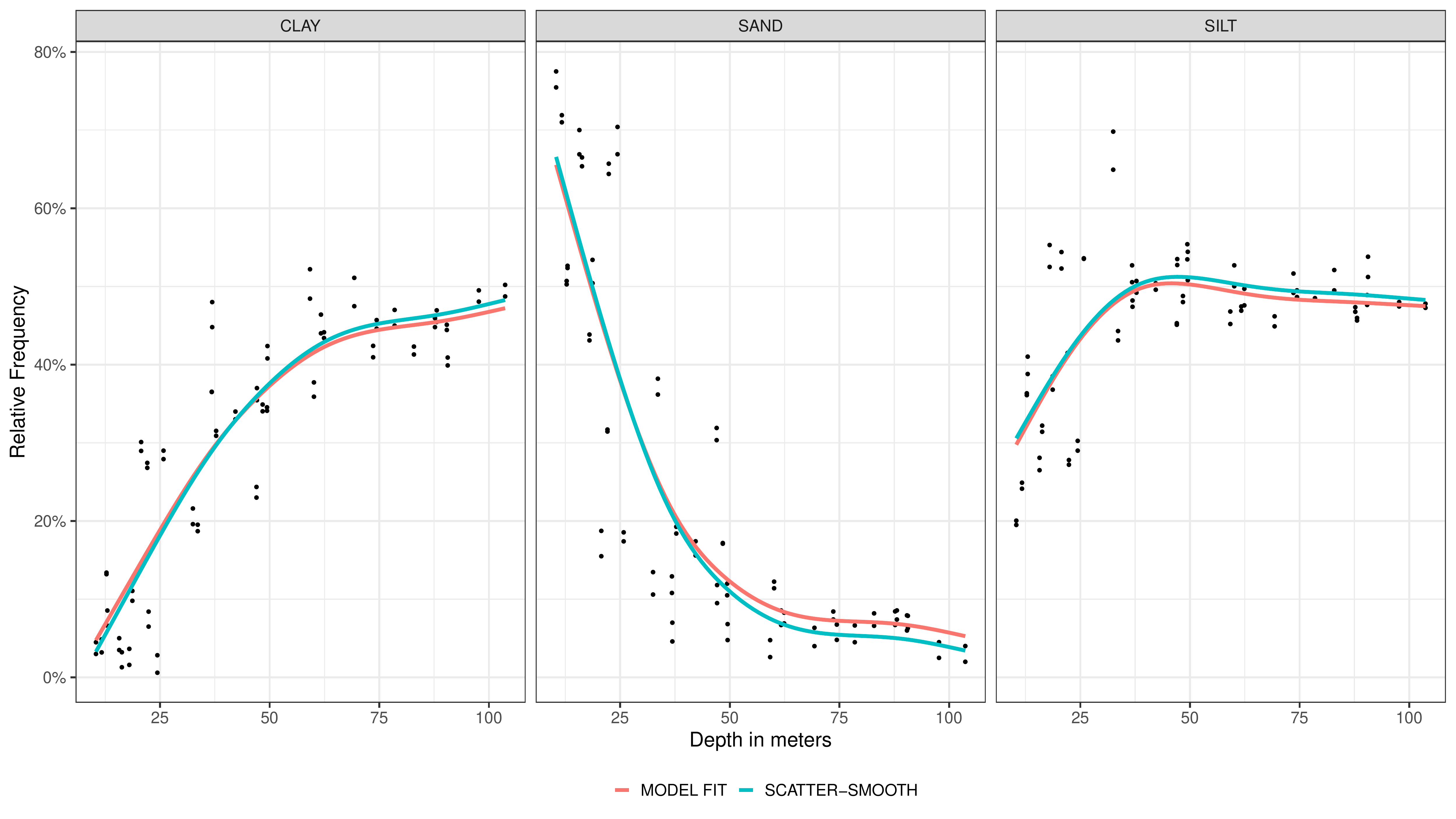}
	\label{fig:arctic_lake_fit}
\end{figure}

\noindent The dataset includes 39 measurements and we model the sediment composition as a function of water depth in meters. Figure \ref{fig:arctic_lake_fit} compares the results of our model to a scatter-smooth estimate. To facilitate visual comparison, we use smoothed estimates of our model.

\noindent The scatter-smooth and our model estimates are in close agreement, showing that with increasing depth, the relative frequency of sand decreases while the proportion of silt and clay increases.

In summary, despite some deviations, our multivariate XGBoostLSS models approximate well the conditional moments of the underlying data generating processes for all distributions studied. For the simulated datasets, one could further improve the accuracy by increasing the trails of the hyper-parameter search or the number of boosting iterations, which we set to 100.

\subsection{Regression Datasets} \label{sec:reg_evaluation} 

We benchmark our model, which we refer to as mXGBoostLSS, against the multivariate NGBoost (mNGBoost) model of \cite{OMalley.2021} and our univariate XGBoostLSS implementation \citep{Marz.2019} using a subset of the datasets of \cite{SpyromitrosXioufis.2016}, as well as the California housing dataset of \cite{Pace.1997}:\footnote{We restrict the datasets in \cite{SpyromitrosXioufis.2016} to continuous or close-to continuous responses only. We use the California housing dataset of scikit-learn. All other datasets are publicly available at \url{http://mulan.sourceforge.net/datasets-mtr.html}. We run all experiments on a 8-core Intel(R) i7-7700T CPU with 32 GB of RAM.}

\begin{itemize}
	\setlength\itemsep{0.5em}
	
	\item \emph{Airline Ticket Price (1d)}: This dataset is concerned with the prediction of airline ticket prices and the target variables are the next day price for 6 flight preferences: (1) any airline with any number of stops, (2) any airline non-stop only, (3) Delta
	Airlines, (4) Continental Airlines, (5) Airtrain Airlines, and (6) United Airlines. 
	
	\item \emph{California Housing}: This dataset was derived from the 1990 U.S. census and contains information with respect to demography, location, as well as general information regarding the house in Californian districts. As responses, we use the median income and the median house value. 
	
	\item \emph{Jura}: The dataset consists of measurements of concentrations of seven heavy metals recorded at 359 locations in the topsoil of a region of the Swiss Jura. The type of land use and rock type are also recorded for each location. Cadmium, copper and lead are treated as targets, while the remaining metals along with land use type, rock type and the coordinates of each location are used as features.
	
	\item \emph{Occupational Employment Survey (2010)}: The Occupational Employment Survey data were obtained from the annual Occupational Employment Survey compiled by the US Bureau of Labor Statistics in 2010. Each target denotes the estimated number of full-time equivalent employees across many employment types (e.g., doctor, dentist, car repair technician, etc.) across 403 cities.
	
	\item \emph{River Flow 1}: The dataset, obtained from the US National Weather Service and collected between September 2011 and September 2012, contains data from hourly river flow observations for 8 sites in the Mississippi River network and the task is to predict river network flows at specific locations.	
		
	\item \emph{Supply Chain Management (1d)}: This dataset is derived from the Trading Agent Competition in
	Supply Chain Management (TAC SCM) tournament from 2010 and contains 16 targets, each corresponding to the next day mean price for each product in the competition. 	
	
	\item \emph{Slump}: This dataset is concerned with the prediction of three properties of concrete (slump, flow and compressive strength) as a function of the content of seven concrete ingredients.
\end{itemize}

\noindent We also include a subset of the simulated trivariate Gaussian and Student-T datasets presented in Section \ref{sec:simulation}, with additional noise features added. Table \ref{tab:data_overview} presents the data and its characteristics. 

\begin{table}[h!]
	\begin{center}
		\begin{threeparttable}
			\caption{Dataset Overview}
			\begin{tabular}{lccccr}
			\toprule
			Dataset                               & Abbreviation       & Observations & $Y_{D}$ & Features & Dependency-Strength \\   
			\midrule
			Airline Ticket Price (1d)             & atp1d              &  337         & 6   & 411   & 0.8013 [0.7305, 0.9166]  \\ 
			California Housing                    & ch                 &  20,640      & 2   & 7     & 0.6881 [0.6881, 0.6881]  \\ 
			Jura                                  & ju                 &  359         & 3   & 15    & 0.1907 [0.1567, 0.2452]  \\ 
			Occupational Employment Survey (2010) & oes10              &  403         & 16  & 298   & 0.549 [0.3195, 0.6928]   \\     
			River Flow 1                          & rf1                &  9,125       & 7   & 65    & 0.5028 [0.0799, 0.8208]  \\   
			Supply Chain Management (1d)          & scm1d              &  9,803       & 16  & 280   & 0.6256 [0.4857, 0.8386]  \\     
			Simulated Trivariate Gaussian         & stg                &  2,000       & 3   & 5     & 0.4585 [0.4542, 0.5035]  \\   
			Simulated Trivariate Student-T        & stt                &  3,000       & 3   & 6     & 0.5524 [0.5261, 0.6780]   \\  
			Slump                                 & sl                 &  103         & 3   & 7     & -0.124 [-0.2035, 0.7001] \\    				
			\bottomrule
			\end{tabular}
			\begin{tablenotes}
				\scriptsize
				\item \hspace{-0.7em} Due to its high level of skewness that caused instabilities for all models, we removed one target (target\_NASI2\_48H\_0) from the rf1 dataset. Further, also for stability reasons, we applied a Box-Cox transformation to all responses of the oes10 dataset. The last column measures the median strength of dependency between responses using the Pearson coefficient of correlation, with additional quantiles in parentheses, i.e., $q_{0.5}(q_{0.1}, q_{0.9})$.
			\end{tablenotes}
			\label{tab:data_overview}
	\end{threeparttable}
	\end{center}
\end{table}

\noindent From Table \ref{tab:data_overview} it follows that there is a non-negligible dependence between targets in all datasets. It is interesting to see how the univariate baseline, which assumes independence between targets, compares to the multivariate models. For conducting the experiments, we create 11 randomly shuffled folds for all datasets. Each is split into train and test, where 80\% is used for training and the remaining 20\% for evaluation. The first fold is used for hyper-parameter tuning only, where we run 100 hyper-parameter trails for each model-dataset combination and select the best set of hyper-parameters. All models are initialized with 500 boosting rounds, with the optimal number of iterations based on early-stopping. Once optimized, we keep the optimal set of hyper-parameters constant and use the remaining 10 folds for model evaluation.\footnote{For the high-dimensional scm1d and oes10-datasets, we were not able to complete the hyper-parameter search for mNGBoost, mainly due to out-of-memory issues, non-positive definite covariance matrices, or unrealistically long runtimes for a single iteration. For these reasons, we had to re-initialize the hyper-parameter search several times and also reduce the number of hyper-parameter trails to 20.} All models except the Student-T assume a Gaussian distribution, with the accuracy being evaluated using the negative log-likelihood (NLL). Table \ref{tab:nll_score} reports the results.

\begin{table}[h!]
\begin{center}
\scalebox{0.8}{
	\begin{threeparttable}
		\caption{NLL scores}
		\begin{tabular}{lrrrrrrr}
			\toprule
			               & mNGBoost-G-C    & mXGBoostLSS-G-C & mXGBoostLSS-G-LRA(5) &  mXGBoostLSS-T-C &  uXGBoostLSS-G \\ 
			\midrule
			atp1d          & \textbf{32.1131 [31.1064, 33.3593]} & 33.1507 [32.4934, 33.8251] &   35.2945 [34.7998, 37.9187] & 34.2416 
			               [32.8878, 35.8722] &    36.4295 [35.6846, 37.9606] \\
			ch             & 1.5875 [1.5602, 1.656] &    1.6921 [1.6512, 1.7364] &      2.2747 [2.2289, 2.3244] &    1.6679 [1.6375, 
			               1.7526] &  \textbf{1.5608 [1.5369, 1.6244]} \\
			ju             &  6.9913 [6.054, 7.6863] &   \textbf{6.9372 [6.6957, 7.489]} &      7.0017 [6.5767, 7.1895] &    7.4975 
			               [7.1694, 7.8141] &        7.5602 [7.1841, 8.068] \\
			oes10          &   4.9139 [3.8811, 5.8206] &     5.763 [5.3078, 6.6205] &  9.1977 [8.3772, 10.0111] &    \textbf{4.0237 
			               [3.5981, 5.3306]} &  4.4383 [3.798, 6.6723] \\
			rf1            & 26.5513 [26.3026, 26.8497] & 21.3728 [21.2832, 21.4263] &    24.9533 [24.337, 25.1423] & 23.3522 [23.1668,  
			               24.2478] &   \textbf{19.1992 [19.0538, 20.0179]} \\
			scm1d          & 97.5695 [97.3239, 97.9193] & 95.5774 [95.4184, 95.8474] & 128.814 [128.1463, 129.0811] &   \textbf{95.124 
				           [95.014, 95.3738]} & 107.1911 [107.0608, 107.3768] \\
			stg            & 3.706 [3.5851, 3.8114] &     \textbf{3.664 [3.5365, 3.7215]} &  3.9296 [3.7848, 4.0291] &     3.8374 [3.7553, 
			               3.898] &       4.6199 [4.5147, 4.7021] \\                
			stt            & 5.1247 [4.7432, 5.6858] &    4.8221 [4.7306, 5.7158] &      5.1441 [4.8861, 5.4495] &    \textbf{4.4243  
				           [4.3407, 4.4598]} &       5.7544 [5.5818, 6.8563] \\
			sl             &   \textbf{10.0841 [9.4764, 10.9282]} &  10.4098 [10.1858, 11.065] &   10.7407 [10.5641, 11.2718] &  10.6563 
			               [10.3614, 11.573] &    12.0785 [11.4485, 13.5503] \\
			\midrule
			Average Rank   & \multicolumn{1}{c} {2.4} & \multicolumn{1}{c} {\textbf{2.2}} & \multicolumn{1}{c} {4.2} & \multicolumn{1}{c} {2.4} & \multicolumn{1}{c} {3.7} \\ 
			\bottomrule
		\end{tabular}
		\begin{tablenotes}
			\scriptsize
			\item \hspace{-0.7em} The table shows median NLL scores across models, datasets and folds, with additional quantiles in parentheses, i.e., $q_{0.5}(q_{0.1}, q_{0.9})$. Lower is better, with best results in bold. At the bottom, we also report average ranks across datasets. Again, lower is better. The columns are to be read as follows: Model-Distribution-Covariance Approximation, where G: Gaussian, T: Student-T, C: Cholesky, LRA: Low-Rank Approximation($r$), m: multivariate, u: univariate. 
		\end{tablenotes}
		\label{tab:nll_score}
\end{threeparttable}
}
\end{center}
\end{table}

\noindent Table \ref{tab:nll_score} shows that our models compare well with existing implementations. With an average rank of 2.2, mXGBoostLSS-G-C has the highest overall accuracy, closely followed by mXGBoostLSS-T-C and mNGBoost-G-C with an average rank of 2.4 each. For kurtotic datasets, the mXGBoostLSS-T-C outperforms its Gaussian counterparts due to its additional degrees of freedom parameter. Probably due to its fixed and non-optimized rank parameter $r$, the LRA model ranks last in our comparison. For the comparatively high-dimensional scm1d and oes10-datasets, the LRA model shows some diverging behaviour. We will further investigate the effect of the rank on the LRA model in Section \ref{sec:ablation}. The comparison between models also shows that explicitly modelling dependencies between targets tends to increase accuracy: for 7 out of 9 datasets, the multivariate models have a higher accuracy than the univariate model. This is consistent with the results of \cite{Schmid.2023} who report that the performances of multivariate approaches were substantially better than the univariate ones for some of the simulation settings considered. 

To further benchmark our models, Table \ref{tab:nll_var} reports the variability ($NLL_{max} - NLL_{min}$) of the NLL-scores across folds and datasets, demonstrating that explicit modelling of dependencies tends to stabilize estimation.

\begin{table}[h!]
	\begin{center}
		\begin{threeparttable}
			\caption{NLL Variability}
			\begin{tabular}{lrrrrrrr}
				\toprule
				& mNGBoost-G-C    & mXGBoostLSS-G-C & mXGBoostLSS-G-LRA(5) &  mXGBoostLSS-T-C &  uXGBoostLSS-G \\ 
				\midrule
			 atp1d &   2.6763 &  \textbf{1.8471} &   5.4083 &     3.8626 &  3.3171 \\
			 ch    &   0.1212 &  0.1120 &   0.1290 &     0.1227 &  \textbf{0.1106} \\
			 ju    &   2.2576 &  1.2288 &   1.0296 &     \textbf{0.6547} &  1.0135 \\
			 oes10 &   2.6402 &  \textbf{1.7893} &   2.6149 &     2.3389 &  3.8587 \\
			 rf1   &   1.3828 &  \textbf{0.3413} &   1.1937 &     2.4426 &  1.1768 \\
			 scm1d &   0.8980 &  0.6472 &   1.4308 &     0.6590 &  \textbf{0.5307} \\
			 stg   &   0.3263 &  0.2021 &   0.3171 &     \textbf{0.1691} &  0.3044 \\
			 stt   &   1.7744 &  1.0664 &   1.0759 &     \textbf{0.2368} &  1.5770 \\
			 sl    &   1.6586 &  1.2184 &   \textbf{0.9302} &     1.3211 &  12.1062 \\
				\midrule
				Average   & 1.5262 &                   \textbf{0.9392} &               1.5699 &                   1.3119 &                 2.6661 \\
				\bottomrule
			\end{tabular}
			\begin{tablenotes}
				\scriptsize
				\item \hspace{-0.7em} The table shows the distance between maximum and minimum NLL scores ($NLL_{max} - NLL_{min}$) across models, dataset and folds. Lower is better, with best results in bold. At the bottom, we also report average distances across datasets. Again, lower is better. The columns are to be read as follows: Model-Distribution-Covariance Approximation, where G: Gaussian, T: Student-T, C: Cholesky, LRA: Low-Rank Approximation($r$), m: multivariate, u: univariate. 
			\end{tablenotes}
			\label{tab:nll_var}
		\end{threeparttable}
	\end{center}
\end{table} 

\noindent Compared to its multivariate counterparts, the univariate model shows the highest average variability across datasets and folds. As an example, consider the sl-dataset, for which the univariate XGBoostLSS model shows some divergent predictions. The lower overall variability might be attributed to the fact that joint modelling of all targets tends to stabilize the estimation in the multivariate setting. Within the class of multivariate models, mXGBoostLSS-G-C shows the least variation, closely followed by mXGBoostLSS-T-C.

In addition to assessing the accuracy, Table \ref{tab:runtime} presents an overview of normalized runtimes. To ensure a fair comparison, we set all hyper-parameters of the models to the same values. All models are estimated using CPUs and all XGBoostLSS models, both univariate and multivariate, are trained without leveraging its fast histogram tree-growing method. We exclude the univariate XGBoostLSS model from the analysis since it has less parameters compared to its multivariate counterparts and therefore always lower runtimes. 

\begin{table}[h!]
	\begin{center}
		\begin{threeparttable}
			\caption{Relative Median Runtimes}
			\begin{tabular}{lccccccc}
				\toprule
				               &  mNGBoost-G-C  & mXGBoostLSS-G-C & mXGBoostLSS-G-LRA(5) &  mXGBoostLSS-T-C &  \\ 
				\midrule
				atp1d          &  5.4236 &                   \textbf{1.0000} &               2.1106 &                   1.0881 \\
				ch             &  6.5707 &                   \textbf{1.0000} &              10.8340 &                   1.4833 \\
				ju             &  1.2184 &                   1.0733 &               3.8502 &                   \textbf{1.0000} \\
				oes10          &  5.5625 &                   1.1406 &               \textbf{1.0000} &                   1.4297 \\                    
				rf1            &  10.0112 &                  1.0413 &               2.6890 &                   \textbf{1.0000} \\
				scm1d          &  45.1237 &                      1.1741 &               2.0985 &                   \textbf{1.0000} \\
				stg            &  3.1539 &                   1.0352 &               7.4486 &                   \textbf{1.0000} \\
				stt            &  3.5399 &                   \textbf{1.0000} &               9.0137 &                   1.3316 \\
				sl             &  \textbf{1.0000} &                   2.0964 &               8.9662 &                   3.0045 \\
				\midrule
				Average Rank   & 3.2           & \textbf{1.7}       & 3.3                    & 1.8 \\ 	
				\bottomrule
			\end{tabular}
			\begin{tablenotes}
				\scriptsize
				\item \hspace{-0.7em} The table shows relative median runtimes, with entries normalized to the model with the lowest runtime. Lower is better, with best results in bold. The following hyper-parameters are used: learning-rate=0.1, max-depth=6, iterations=100. All other hyper-parameters are set to their default values. The columns are to be read as follows: Model-Distribution-Covariance Approximation, where G: Gaussian, T: Student-T, C: Cholesky, LRA: Low-Rank Approximation($r$), m: multivariate, u: univariate. 
			\end{tablenotes}
			\label{tab:runtime}
		\end{threeparttable}
	\end{center}
\end{table}

\noindent From Table \ref{tab:runtime} it follows that, while mNGBoost-G-C has the lowest runtime for the smallest sl-dataset, the Cholesky-based mXGBoostLSS models scale well with the number of observations and with the mXGBoostLSS-G-C model being the most efficient in terms of runtime across datasets. The efficiency is likely to increase further if we leverage XGBoostLSS's GPU-histogram training. Also, for fairly large datasets, one can use distributed training with Dask for even better scalability. Table \ref{tab:runtime} also shows that the LRA-model benefits from its linear increase in parameters which results in the lowest runtime for the high-dimensional oes10-dataset.

\subsection{Ablation Study} \label{sec:ablation}

Following the discussion in the previous section, it remains to be investigated why the LRA-model does not perform as well as the other models, especially for datasets with small $D$ and $N$. One reason might be the low number of observations relative to the number of estimated parameters. Recall that for the Gaussian, the Cholesky factorization of the covariance matrix requires estimation of $D(D+3)/2$ parameters, while $D(2+r)$ need to be estimated for the low-rank covariance approximation. As an example, take the relatively small sl-dataset with $D=3$: while only 9 parameters have to be estimated for the Cholesky model, there are already 21 parameters for the LRA-model with $r=5$. 

Another reason might be related to the choice of the rank parameter $r$. While all hyper-parameters of the Cholesky models are optimized, we set $r=5$ for all datasets, mainly to keep the computational cost low while still maintaining a reasonable fit. To further investigate the effect of $r$ on the accuracy, we run additional experiments using a small subset of the datasets with varying values of $r$. Table \ref{tab:ablation} reports the results.

\begin{table}[h!]
	\begin{center}
		\begin{threeparttable}
			\caption{Ablation Results of the LRA($r$) model}
			\begin{tabular}{lrrr}
				\toprule
				Rank ($r$)                                 & atp1d       & ju    & sl \\   
				\midrule
				 2 & \textbf{34.9707 [34.3182, 35.9444]} & 7.2583 [6.9075, 7.6592] & 14.9027 [14.6824, 16.4902] \\
				4 &     35.0910 [34.3140, 36.8900] & 7.5869 [6.9829, 7.9232] & 12.8737 [12.2232, 12.9751] \\
				5 & 35.2945 [34.7998, 37.9187] & 7.0017 [6.5767, 7.1895] & \textbf{10.7407 [10.5641, 11.2718]} \\
				6 & 48.0852 [47.1251, 48.4539] & 8.1454 [7.8824, 8.3371] & 16.8487 [14.7109, 23.4578] \\
				8 & 37.8567 [37.1523, 38.7453] & \textbf{6.9445 [6.2847, 7.4682]} &  13.4723 [13.2427, 13.6850] \\
				10 & 70.3046 [48.6155, 79.8918] & 7.2275 [6.8551, 7.5264] & 14.7997 [14.4707, 18.2046] \\
				\bottomrule
			\end{tabular}
			\begin{tablenotes}
				\scriptsize
				\item \hspace{-0.7em} The table shows median NLL scores across datasets, folds and varying values of $r$ for the LRA model, with additional quantiles in parentheses, i.e., $q_{0.5}(q_{0.1}, q_{0.9})$. Lower is better, with best results in bold.
			\end{tablenotes}
			\label{tab:ablation}
		\end{threeparttable}
	\end{center}
\end{table}

\noindent Table \ref{tab:ablation} shows that the quality of the covariance approximation varies with the rank parameter. However, there is no general recommendation that higher values of $r$ tend to increase accuracy. This can be seen from the atp1d-dataset, where the model tends to overfit with increasing values of $r$, resulting in lower accuracy. For the relatively low-dimensional ju-dataset, higher values of $r$ increase accuracy, while for the sl-dataset, a moderately low rank gives the best results. Since the order of $r$ is depending on the size of the dataset and the characteristics determining the covariance structure, our recommendation would be to add the rank parameter as a tuneable hyper-parameter.

\section{Conclusion, Limitations and Future Research} \label{sec:conclusion}

While most implementations of Gradient Boosting Machines are tailored to single-target settings, this paper presents an extension of XGBoostLSS that models multi-targets and their dependencies in a probabilistic regression environment. Using simulation studies and real-world data, we have shown that our approach outperforms existing GBMs with respect to runtime and is competitive in terms of accuracy. We have also demonstrated that explicit modelling of dependencies between targets can lead to an increase in accuracy.

Despite its flexibility, we acknowledge some limitations of our approach that require additional research. Although the base XGBoost implementation accepts a $N \times D$ array of responses, model training is still optimized for single-target models. This implies that for our XGBoostLSS approach, which is based on multi-parameter training, with a separate tree grown for each parameter, estimating many parameters for a large dataset can become computationally expensive, with the computational cost growing $\mathcal{O}(K^{2})$. We would like to emphasize, though, that high-dimensional multi-target regression and multiclass-classification is a known scaling problem of current GBMs implementations and therefore not a unique limitation of our approach. An interesting scope for future implementation and research would be a more runtime efficient version of our framework, where multiple parameters can be estimated with a single tree. In addition, we consider the extension of our framework to allow for a more flexible choice of multivariate response distributions beyond the Gaussian, Dirichlet and Student-T to be an interesting refinement.

\bibliography{literature}


\end{document}